\documentclass[11pt,a4paper]{article}

\usepackage[hyperref]{acl2021}
\usepackage{times}
\usepackage{amsmath}
\usepackage{amsfonts}
\usepackage{natbib}
\usepackage{graphicx}
\usepackage{latexsym}
\usepackage{multirow}
\usepackage{cleveref}
\usepackage[ruled,linesnumbered]{algorithm2e}

\usepackage{algpseudocode}  
\author{ \parbox{\linewidth}{\centering Wenhao Wu\textsuperscript{1}\thanks{\ \ Work is done during an internship at Baidu Inc.}, Wei Li\textsuperscript{2}, Xinyan Xiao\textsuperscript{2}, Jiachen Liu\textsuperscript{2},Ziqiang Cao\textsuperscript{3}, Sujian Li\textsuperscript{1}\thanks{\ \ Corresponding author.}, Hua Wu\textsuperscript{2}, Haifeng Wang\textsuperscript{2} }\\
    \textsuperscript{1}Key Laboratory of Computational Linguistics, MOE, Peking University \\
  \textsuperscript{2}Baidu Inc., Beijing, China \\
  \textsuperscript{3}Institute of Artificial Intelligence, Soochow University, China\\
  \texttt{\{waynewu,lisujian\}@pku.edu.cn}\\
  \texttt{\{liwei85,xiaoxinyan,liujiachen,wu\_hua,wanghaifeng\}@baidu.com}\\
  \texttt{\{zqcao\}@suda.edu.cn}\\

  }

\usepackage{microtype}

\aclfinalcopy 
\def\aclpaperid{2198} 

\title{BASS: Boosting Abstractive Summarization with  Unified Semantic Graph}

\date{}

\begin{document}
\maketitle

\begin{abstract}
 
Abstractive summarization for long-document or multi-document remains challenging for the  Seq2Seq architecture, as Seq2Seq is not good at analyzing long-distance relations in text.
In this paper, we present BASS, a novel framework for Boosting Abstractive Summarization based on a unified Semantic graph, 
which aggregates co-referent phrases distributing across a long range of 
context and conveys rich relations between phrases.
Further, a graph-based encoder-decoder model is  proposed to improve both the document representation and summary generation process by leveraging the graph structure.
Specifically, several graph augmentation methods are designed to encode both the explicit and implicit relations in the text while the graph-propagation attention mechanism is developed in the decoder to select salient content into the summary.
Empirical results show that the proposed architecture brings substantial improvements for both long-document and multi-document summarization tasks.

\end{abstract}

\section{Introduction}


Nowadays, the sequence-to-sequence (Seq2Seq) based summarization models  have gained unprecedented popularity \citep{rush-etal-2015-neural,
see-etal-2017-get,lewis-etal-2020-bart}.
However, complex summarization scenarios such as long-document  or multi-document summarization (MDS), still bring great challenges to Seq2Seq models \cite{cohan-etal-2018-discourse,DBLP:conf/iclr/LiuSPGSKS18}.
In a long document numerous details and salient content may distribute evenly \cite{sharma-etal-2019-bigpatent} while multiple documents may contain repeated, redundant  or contradictory information \cite{radev-2000-common}. These problems make Seq2Seq models struggle with content selection and organization which mainly depend on the long source sequence \cite{DBLP:journals/corr/ShaoGBGSK17}. 
Thus, how to  exploit deep semantic structure in the complex text input is a key to further promote summarization performance.
\begin{figure}
    \includegraphics[scale=0.30]{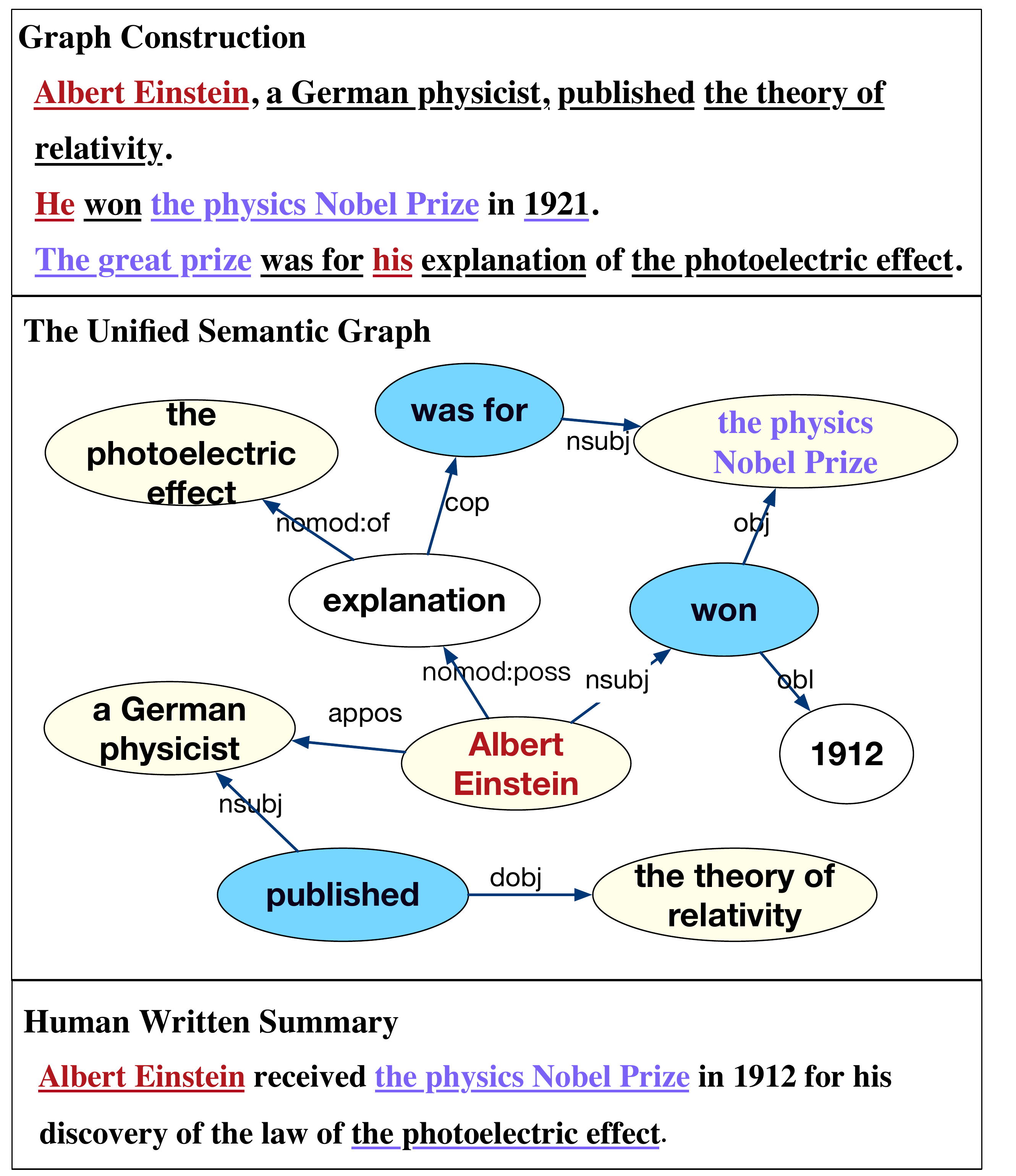}
    \caption{Illustration of a unified semantic graph and its construction procedure for a document containing three sentences. In \textbf{Graph Construction}, underlined tokens represent phrases., co-referent phrases are represented in the same color. In \textbf{The Unified Semantic Graph}, nodes of different colors indicate different  types, according to section \ref{sec:graph}.}
    \label{fig:graph}
\end{figure}

Compared with sequence, graph can aggregate relevant disjoint context by uniformly representing them as nodes and their relations as edges.
This greatly benefits  global structure learning and  long-distance relation modeling.
Several previous works have attempted to leverage sentence-relation graph to improve long sequence summarization, where nodes are sentences and edges are similarity or discourse relations between sentences \cite{li-etal-2020-leveraging}.
However, the sentence-relation graph is not flexible for fine-grained (such as entities) information aggregation and relation modeling.
Some other works also proposed to construct  local knowledge graph by OpenIE to improve Seq2Seq models \cite{fan-etal-2019-using,huang-etal-2020-knowledge}.
However, the OpenIE-based graph only contains sparse relations between partially extracted phrases, which cannot reflect the global structure and rich relations of the overall sequence.

For better modeling the long-distance relations and global structure of a long sequence, we propose to apply a phrase-level unified semantic graph to facilitate content selection and organization.
Based on fine-grained phrases extracted from dependency parsing, our graph is suitable for information aggregation with the help of coreference resolution that substantially compresses the input and benefits content selection.
Furthermore, relations between phrases play an important role in organizing the salient content when generating summaries. 
For example, in Figure \ref{fig:graph} the phrases ``Albert Einstein'', ``the great prize'' and ``explanation of the of the photoelectric''  which distribute in different sentences are easily aggregated through their semantic relations to compose the final summary sentence.

We further propose a graph-based encoder-decoder model based on the unified semantic graph.
The graph-encoder effectively encodes long sequences by explicitly modeling the relations between phrases and capturing the global structure based on the semantic graph.
Besides, several graph augmentation methods are also applied during graph encoding to tap the potential semantic relations.
For the decoding procedure, the graph decoder incorporates the graph structure by graph propagate attention to guide the summary generation process, which can help select salient content and organize them into a coherent summary.

We conduct extensive experiments on both the long-document summarization dataset BIGPATENT and MDS dataset WikiSUM to validate the effectiveness of our model.
Experiment results demonstrate that our graph-based model significantly improves the performance of both long-document and multi-document summarization over several strong baselines.
Our main contributions are summarized as follows:
\begin{itemize}
\item We present the unified semantic graph which aggregates co-referent phrases distributed in context for better modeling the long-distance relations and global structure in long-document summarization and MDS. 
\item We propose a  graph-based encoder-decoder model to improve both the document representation and summary generation process of the Seq2Seq architecture by leveraging the graph structure. 
\item Automatic and human evaluation on both long-document summarization and MDS outperform several strong baselines and validate the effectiveness of our graph-based model.
\end{itemize}

\section{Related Works}
\subsection{Abstractive  Summarization}
Abstractive summarization aims to generate a fluent and concise summary for the given input document \citep{rush-etal-2015-neural}.
Most works apply Seq2Seq architecture to implicitly learn the summarization procedure \citep{see-etal-2017-get,gehrmann-etal-2018-bottom,DBLP:journals/corr/PaulusXS17,celikyilmaz-etal-2018-deep}.
More recently, significant improvements have been achieved by applying pre-trained language models as encoder \cite{liu-lapata-2019-text,rothe-etal-2020-leveraging}  or pre-training the generation process leveraging a large-scale of unlabeled corpus \citep{DBLP:conf/nips/00040WWLWGZH19, lewis-etal-2020-bart,qi-etal-2020-prophetnet,DBLP:conf/icml/ZhangZSL20}. 
In MDS, most of the previous models apply extractive methods \cite{erkan2004lexrank,cho-etal-2019-multi}. 
Due to the lack of large-scale datasets, some attempts on abstractive methods transfer single document summarization (SDS) models to MDS \cite{lebanoff-etal-2018-adapting,
10.1145/3357384.3357872} or unsupervised  methods based on auto-encoder \citep{DBLP:conf/icml/ChuL19,brazinskas-etal-2020-unsupervised,amplayo-lapata-2020-unsupervised}.
After the release of several large MDS datasets \cite{DBLP:conf/iclr/LiuSPGSKS18,fabbri-etal-2019-multi}, some supervised abstractive models for MDS appear \cite{liu-lapata-2019-hierarchical,li-etal-2020-leveraging}.
Their works also emphasize the importance of modeling cross-document relations in MDS.
\subsection{Structure Enhanced Summarization}
Explicit structures play an important role in recent deep learning-based extractive and abstractive summarization methods \cite{DBLP:conf/emnlp/LiXLW18,DBLP:conf/emnlp/LiXLW18a,DBLP:conf/naacl/LiuTL19}.
Different structures benefit summarization models from different aspects.
Constituency parsing greatly benefits content selection and compression for extractive models.
\citet{10.5555/2886521.2886620} propose to extract salient sentences based on their constituency parsing trees. \citet{xu-durrett-2019-neural}  and \citet{desai-etal-2020-compressive} jointly select and compress salient content based on  syntax structure and  syntax rules.
Dependency parsing helps summarization models in semantic understanding.
\citet{jin2020semsum}  incorporate semantic dependency graphs of input sentences to help the summarization models generate sentences  with better  semantic relevance .
Besides sentence-level structures, document-level structures also attract a lot of attention. 
\citet{DBLP:conf/iclr/FernandesAB19} build a simple graph consisting of sentences, tokens and POS   for summary generation. 
By incorporating  RST trees, \citet{xu-etal-2020-discourse} propose a discourse-aware model to extract sentences. 
Similarly, structures from semantic analysis also help. 
\citet{liu-etal-2015-toward}  and \citet{liao-etal-2018-abstract} propose to guide summarization with Abstract Meaning Representation (AMR) for a better comprehension of the input  context.
Recently, the local knowledge graph by OpenIE attracts great attention.
Leveraging OpenIE extracted tuples, \citet{fan-etal-2019-using}  compress and reduce redundancy in multi-document inputs in MDS.
Their work mainly focus on the efficiency in processing long sequences.
\citet{huang-etal-2020-knowledge} utilize OpenIE-based graph for boosting the faithfulness of the generated summaries.
Compared with their work, our phrase-level semantic graph focus on modeling long-distance relations and semantic structures. 
\begin{table}
    \centering
    \begin{tabular}{l|c|c|c|c}
         \hline
         Input  Length&800&1600&2400&3000  \\
         \hline
         \#Nodes&140&291&467&579\\
         \#Edges&154&332&568&703\\
         \hline
    \end{tabular}
    
    \caption{Illustration of 
    how the average number of nodes and edges in the graph changes when the input sequence becomes longer on WikiSUM.}
    \label{tab:graph_stats}
\end{table}
\section{Unified Semantic Graph}
In this section, we introduce the definition and construction of the unified semantic graph.
\subsection{Graph Definition}\label{sec:graph}
The unified semantic graph is a heterogeneous graph defined as $G=(V,E)$, where $V$ and $E$ are the set of nodes and  edges.
Every node in $V$ represents a concept merged from co-referent phrases.
For example, in Figure \ref{fig:graph} the node ``\textit{Albert Einstein}" is merged from phases  ``\textit{Albert Einstein}" and  ``\textit{his}" which indicate the same person by coreference resolution.
Defined as a heterogeneous graph $G$,  every node $v \in V$ and every edge  $e_{ij} \in E$  in our graph belongs to a type of phrase and dependency parsing  relation, respectively.
Determined by the type of phrases merged from, nodes are categorized into three different types: Noun phrase (N), Verb phrase (V), Other phrase (O).
We neglect dependency relations in edges as they  mainly indicate sentence syntax.
Instead, the meta-paths \cite{sun2011pathsim} in the unified semantic graph convey various semantic relations.
Notice that most O such as adjective phrases, adverb phrases function as modifiers, and the meta-path \textbf{ O-N } indicates modification relation.
The  meta-path \textbf{N-N} between Noun phrases represents appositive relation or appositional relation.
Furthermore, two-hop meta-path represents more complex semantic relations in graph.
For example,\textbf{ N-V-N} like [\textit{Albert Einstein}]-[\textit{won}]-[\textit{the physics Nobel Prize}]  indicates  SVO (subject–verb–object) relation.
It is essential to effectively model the two-hop meta-path for complex semantic relation modeling.
\subsection{Graph Construction}\label{sec:merge}
To construct the semantic graph, we extract phrases and their relations from sentences by first merging tokens into phrases and then  merging co-referent phrases into nodes.
We employ CoreNLP \cite{manning-etal-2014-stanford} to obtain coreference chains of the input sequence and the dependency parsing tree of each sentence.
Based on the dependency parsing tree, we merge consecutive tokens that form a complete  semantic unit into a phrase.
Afterwards, we merge the same phrases from different positions and phrases in the same coreference chain to form the nodes in the semantic graph. 

The final statistics of the unified semantic graph on WikiSUM are illustrated in table \ref{tab:graph_stats}, which indicates that the scale of the graph expands moderately with the inputs.
This also demonstrates how the unified semantic graph compresses long-text information.

\section{Summarization Model}
\begin{figure*}
    \includegraphics[scale=0.38]{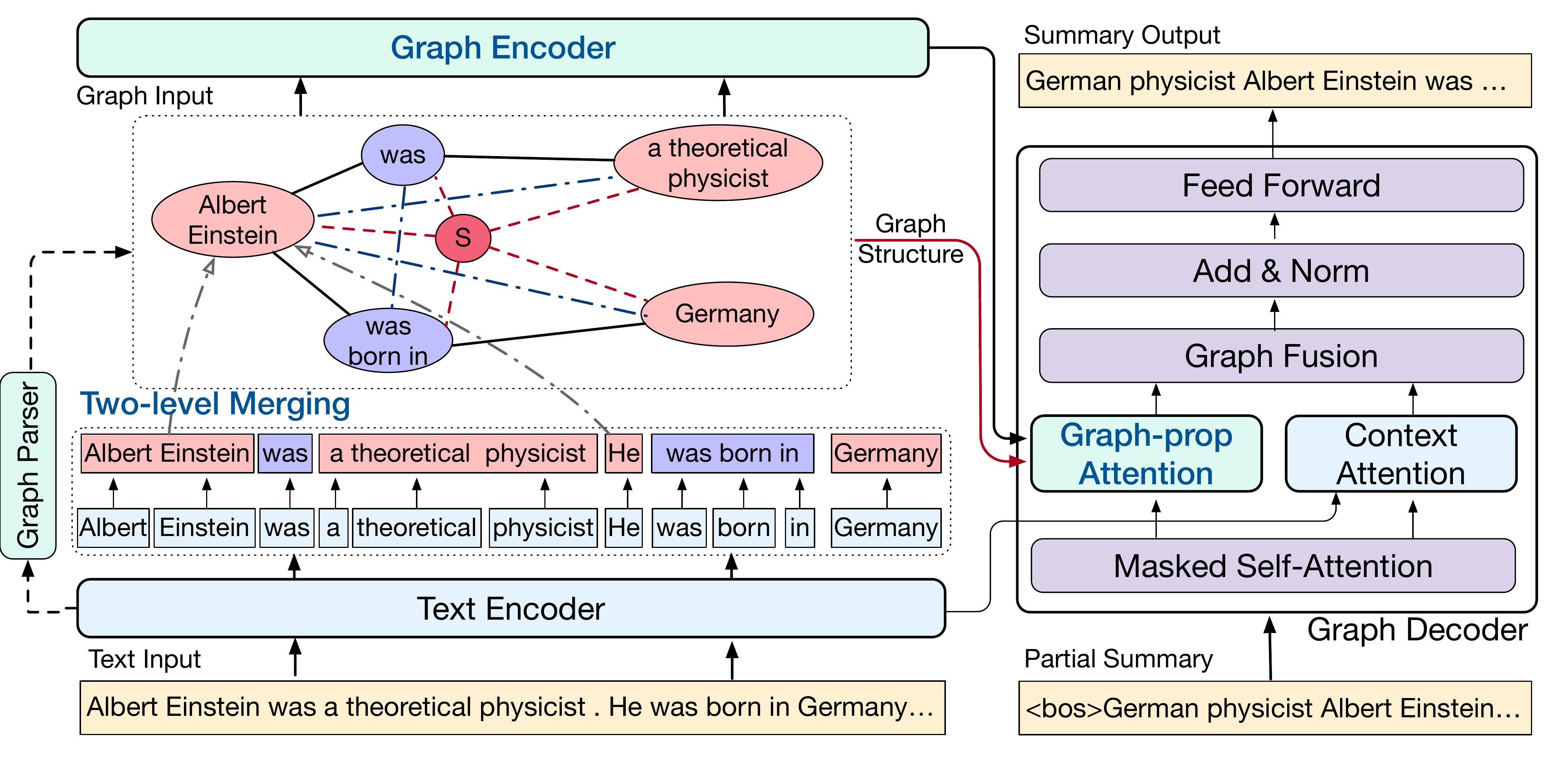}
    \caption{Illustration of our  graph-based summarization model. The graph node representation is  initialized from merging token representations in two-level. The graph encoder models the augmented graph structure. The decoder attends to both token and node representations and utilizes graph structure by graph-propagation attention.}
    \label{fig:mehod}
\end{figure*}

In this section, we introduce our graph-based abstractive summarization model, which mainly consists of a graph encoder and a graph decoder, as shown in Figure \ref{fig:mehod}.
In the encoding stage, our model takes a document or the concatenation of a set of documents as text input (represented as $x=\{x_k\}$), and encodes it by a text encoder to obtain a sequence of local token representations.
The graph encoder further takes the unified semantic graph as graph input (represented as $G=(V, E)$ in \cref{sec:graph}), and explicitly model the semantic relations in graph to obtain global graph representations.
Based on several novel graph-augmentation methods, the graph encoder also effectively taps the implicit semantic relations across the text input.
In the decoding stage, the graph decoder leverages the graph structure to guide the summary generation process by a novel graph-propagate attention, which facilitates salient content selection and organization for generating more informative and coherent summaries.


\subsection{Text Encoder}
To better represent local features in sequence, we apply the pre-trained language model  RoBERTa \cite{liu2019roberta} as our text encoder.
As the maximum positional embedding length of RoBERTa is 512, we extend the positional embedding length and randomly initialize  the  extended  part.
To be specific, in every layer, the representation of every node is only updated by it's neighbors by self attention.

\subsection{Graph Encoder}
After we obtain token representations by the text encoder, we further model the graph structure to obtain node representations.
We initialize node representations in the graph based on token representations and the token-to-node alignment information from graph construction. 
After initialization, we apply graph encoding layers to model the explicit semantic relations features and additionally apply several graph augmentation methods to learn the implicit structure  conveyed by the graph.
 \\
\textbf{Node Initialization} \quad 
Similar to graph construction in  \cref{sec:merge}, we initialize graph representations following the two-level merging, token merging and phrase merging.
The token merging compresses and abstracts local token features into higher-level phrase representations.
The phrase merging aggregates co-referent phrases in a wide context, which captures long-distance and cross-document relations.
To be simple, these two merging steps are implemented by average pooling.
\\
\textbf{Graph Encoding Layer} \quad Following previous works in graph-to-sequence learning \cite{koncel-kedziorski-etal-2019-text,DBLP:conf/acl/YaoWW20}, we apply Transformer layers for graph modeling by  applying the graph adjacent matrix as self-attention mask.
\\
\textbf{Graph Augmentation} \quad Following previous works \cite{bastings-etal-2017-graph,koncel-kedziorski-etal-2019-text}, we add reverse edges and self-loop edges in graph as the original directed edges are not enough for learning backward  information.
For better utilizing the properties of the united semantic graph, we further propose two novel graph augmentation methods.
\\
\textit{Supernode} \quad As the graph becomes larger,  noises introduced by imperfect graph construction also increase, which may cause disconnected sub-graphs.
To strengthen the robustness of graph modeling and learn better global representations, we add a special supernode connected with every other node in the graph to increase the connectivity.\\
\textit{Shortcut Edges}\quad  Indicated by previous works, graph neural networks are weak at modeling multi-hop  relations \cite{pmlr-v97-abu-el-haija19a}.
However, as mentioned in \cref{sec:graph}, the meta-paths of length two represent rich semantic structures that require further modeling the two-hop relations between  nodes.
As illustrated in Figure \ref{fig:mehod}, in a \textbf{N-V-N} meta-path [\textit{Albert Einstein}]-[\textit{was}]-[\textit{a theoretical physicist}],   the relations   [\textit{Albert Einstein}]-[\textit{was}] and  [\textit{was}]-[\textit{a theoretical physicist}] are  obviously less important than the two-hop relation   [\textit{Albert Einstein}]- [\textit{a theoretical physicist}].
Therefore we add shortcut edges between every node and its two-hop relation neighbors, represented as blue edges in Figure \ref{fig:mehod}.
We have also attempted other complex methods such as MixHop \cite{pmlr-v97-abu-el-haija19a}, but we find shortcut edges are more efficient and effective.
The effectiveness of these graph augmentation methods has also been validated in \cref{sec:validate}.
\subsection{Graph Decoding Layer}
Token and node representations benefit summary generation in different aspects.
Token representations are better at capturing local features while graph representations provide global and abstract features.
For leveraging both representations, we apply a stack of Transformer-based graph decoding layers as the decoder which attends to both representations and fuse them for generating summaries.
Let  $y_t^{l-1}$ denotes the representation of  $t$-th summary token output by  ($l-1$)-th  graph decoding layer.
For the graph attention, we apply multi-head attention using $y_t^{l-1}$  as query and  node representations $V=\{v_j\}$ as keys and values:
\begin{equation}
\alpha_{t,j} = \frac{(y_t^{l-1}W_Q)(v_jW_{K})^T}{\sqrt{d_{head}}}  \label{graphsalient}
\end{equation}
where $ W_Q, W_{K} \in \mathbb{R}^{d\times d} $ are  parameter weights, $\alpha_{t,j}$ denote the salient score for node $j$ to $y^{l-1}_t$.
We then calculate the global graph vector $g_t$ as weighted sum over values of nodes:
$g_t  = \sum_{j}Softmax(\alpha_{t,j})(v_jW_V)$
where  $W_V \in \mathbb{R}^{d\times d}$ is a learnable parameter.
We also obtain contextualized text vector $c_t$ similar to the procedure above by calculating multi-head attention between $y_t^{l-1}$ and token representations.
Afterwards, we use a  graph fusion layer which is  a feed-forward neural network to fuse the concatenation of the two features: $d^{l}_t = W_d^T([g_t,c_t])\label{fusion}$, where $W_d$ $\in \mathbb{R}^{2d\times d}$  is the linear transformation parameter and $d_t^l$ is the hybrid representation of tokens and graph.
After layer-norm and feed-forward layer, the  $l$-th graph decoding layer output $y_t^l$ is used as the input of the next layer and also used for generating the $t_{th}$ token  in the final layer.
\\
\textbf{Graph-propagate Attention}\quad
When applying multi-head attention to graph, it only attends to  node representations linearly, neglecting the graph structure.
Inspired by \citet{DBLP:conf/iclr/KlicperaBG19}, we propose the graph-propagate attention  to leverage the graph structure to guide the summary generation process.
By further utilizing semantic structure, the decoder is more efficient in selecting  and organizing  salient content.
Without extra parameters, the  graph-propagation attention can be conveniently applied to the conventional multi-head attention for structure-aware learning. 

Graph-propagate attention consists of two steps: salient score prediction and score propagation.
In the first step, we predict the salient score for every node linearly.
We apply the output of multi-head attention  $\alpha_{t}\in\mathbb{R}^{|v|\times C} $ in Equation \ref{graphsalient}  as salient scores, where $|v|$ is the number of nodes in the graph and $C$ is the number of attention heads.
$C$ is regarded  as $C$ digits or channels of the salient score for every node.  
We then make the salient score structure-aware through score propagation.
Though  PageRank can propagate salient scores over the entire graph,  it leads to over-smoothed scores, as in every summary decoding step  only parts of the content are salient.
Therefore, for each node we only propagate its salient score $p$ times in the graph, aggregating at most $p$-hop relations.
Let $\beta^0_{t} = \alpha_{t} $ denotes the initial salient score predicted in previous step, the salient score after $p$-th propagation is:
\begin{equation}
\beta^p_t = \omega\hat{A}\beta^{p-1}_t + (1-\omega)\beta^0_t
\end{equation}
where $\hat{A}=AD^{-1}$ is a degree-normalized adjacent matrix  of the graph\footnote{Adjacent matrix A contains self-loop and reverse edges.}, and $\omega \in (0,1] $ is the teleport probability which defines the salient score has the probability  $\omega$ to propagate towards the neighbor nodes and $1-\omega$  to restart from initial.
The graph-propagation procedure can also be formulated as:
\begin{equation}
\beta^p_t =(\omega^p\hat{A}^p + (1-\omega)(\sum_{i=0}^{p-1}\omega^i\hat{A}^i))\alpha_{t}
\end{equation}
After $p$  steps of salient score propagation, the graph vector is then calculated by weighted sum of node values: 
\begin{equation}
g^{'}_t = \sum_{j}Softmax(\beta^p_{t,j})(v_jW^V)
\end{equation}

where for the convenience of expression,  the concatenation of multi-head is omitted.  
The output of  fusing $g^{'}_t$ and $c_t$ is then applied to generate the $t_{th}$ summary token  as mentioned before.
\section{Experiment Setup}
In this section, we describe the
datasets of our experiments and  various implementation details.
\subsection{Summarization Datasets}
We evaluate our model on a SDS dataset and an MDS dataset, namely BIGPATENT \cite{sharma-etal-2019-bigpatent} and  WikiSUM \cite{DBLP:conf/iclr/LiuSPGSKS18}.\\
\textbf{BIGPATENT} is a large-scale patent document summarization dataset with an average input of 3572.8 words and a reference with average length of 116.5 words. 
BIGPATENT is a highly abstractive summarization dataset with salient content evenly distributed in the input. 
We follow the standard splits of \citet{sharma-etal-2019-bigpatent} for training, validation, and testing (1,207,222/67,068/67,072).\\
\textbf{WikiSUM} is a large-scale MDS dataset. 
Following \citet{liu-lapata-2019-hierarchical}, we treat the generation of lead Wikipedia sections as an MDS task.
To be specific, we directly utilize the preprocessed results from \citet{liu-lapata-2019-hierarchical}, which split source documents into multiple paragraphs and rank the paragraphs based on their titles to select top-40 paragraphs  as  source input.
The average length of each paragraph and the target summary are 70.1 tokens and 139.4 tokens, respectively.
We concatenate all the paragraphs as the input sequence.
We use the standard splits of \citet{liu-lapata-2019-hierarchical} for training, validation, and testing (1,579,360/38,144/38,205). 
\subsection{Implementation Details}
 We train all the abstractive models by max likelihood estimation with label smoothing  (label smoothing factor 0.1). 
As we fine-tune the pre-trained language model RoBERTa as text encoder, we apply two different Adam optimizers \cite{DBLP:journals/corr/KingmaB14} with  $\beta_1=0.9$ and $\beta_2=0.998$  to train the  pre-trained part and other parts of the model \cite{liu-lapata-2019-text}.
The learning rate and warmup steps are 2e-3 and 20,000 for the pre-trained part and  0.1 and 10,000 for other parts.
As noticed from experiments, when the learning rate is high, graph-based models suffer from unstable training caused by the gradient explosion in the text encoder.
Gradient clipping with a very small maximum gradient norm (0.2 in our work) solves this problem.
All the models are trained for 300,000 steps on BIGPATENT and WikiSUM with 8 GPUs (NVIDIA Tesla V100).
We apply dropout (with the probability of 0.1) before all  linear layers.
In our model, the number of graph-encoder layers and graph-decoder layers are set as 2 and 6, respectively.
The hidden size of both graph encoding and graph decoding layers is 768 in alignment with RoBERTa, and the feed-forward size is 2048 for parameter efficiency.
For graph-propagation attention, the parameter $\omega$ is  0.9, and the propagation steps $p$ is 2.
During decoding, we apply beam search  with beam size 5 and length penalty with factor 0.9. 
Trigram blocking is used to reduce repetitions.
\section{Results}
\subsection{Automatic Evaluation}
\begin{table}
\centering
\begin{tabular}{l|lllc}
\hline
Model&R-1&R-2&R-L&BS\\
\hline
Lead&38.22&16.85&26.89&-\\
LexRank& 36.12&11.67&22.52&-\\
\hline
TransS2S&40.56&25.35&34.73&25.43\\
T-DMCA&40.77&25.60&34.90&-\\
HT&41.53&26.52&35.76&25.62\\
BERTS2S&41.49&25.73&35.59&-\\
RoBERTaS2S&42.05&27.00&36.56&29.13\\
GraphSum&42.99&27.83&37.36&29.69\\
\hline
BASS(2400)&43.65&\textbf{28.55}&37.85& \textbf{31.91}\\
BASS(3000)&\textbf{44.33}&28.38&\textbf{37.87}&31.71\\
\hline
\end{tabular}
\caption{Evaluation results on the test set of  WikiSUM. Rouge-1, Rouge-2, Rouge-L and BERTScore are abbreviated as R-1,R-2,R-L and BS, respectively.}
\label{wikisum_exp}
\end{table}
\begin{table}
\centering
\begin{tabular}{l|llcc}
\hline
Model&R-1&R-2&R-L&BS\\
\hline
Lead&31.27&8.75&26.18&-\\
ORACLE&43.56&16.91&36.52&-\\
LexRank& 35.99&11.14&29.60&-\\
\hline
Seq2Seq& 28.74&7.87&24.66&-\\
Pointer&30.59&10.01&25.65&-\\
Pointer+cov&33.14&11.63&28.55&-\\
FastAbs&  37.12& 11.87& 32.45&-\\
TLM &36.41& 11.38&30.88&-\\
TransS2S&34.93&9.86&29.92&9.42\\
 RoBERTaS2S&43.62&18.62&37.86&18.18\\
 BART&\textbf{45.83}&19.53&-&-\\
 Pegasus-base&43.55&\textbf{20.43}&-&-\\
\hline
BASS&45.04&20.32&\textbf{39.21}&\textbf{20.13}\\
\hline
\end{tabular}
\caption{ Evaluation results on the test set of  BIGPATENT where the length input of BASS is 1024.}
\label{table:exp1}
\end{table}
We evaluate the quality of generated summaries using ROUGE  $F_1$\cite{lin2004rouge} and BERTScore \cite{DBLP:conf/iclr/ZhangKWWA20}.
For ROUGE, we report unigram and bigram overlap between system summaries and reference summaries (ROUGE-1, ROUGE-2).
We report sentence-level ROUGE-L for the BIGPATENT  dataset and summary-level ROUGE-L for the WikiSUM for a fair comparison with previous works.
We also report BERTScore \footnote{We apply roberta-large\_L17\_no-idf\_version as the metric model and  rescale\_with\_baseline setting according to suggestions on https://github.com/Tiiiger/bert\_score.} $F_{1}$, a better metric at evaluating semantic similarity  between  system summaries and reference summaries.\\
\textbf{Results on MDS}\quad Table \ref{wikisum_exp} summarizes the evaluation results on the WikiSUM dataset.
We compare our model with several strong abstractive and extractive baselines.
As listed in the top block, Lead and LexRank \cite{erkan2004lexrank} are two classic extractive methods.
The second block shows the results of several different abstractive methods.
TransS2S is the Transformer-based encoder-decoder model.
By replacing the Transformer encoder in TransS2S with BERT \cite{DBLP:conf/naacl/DevlinCLT19} or RoBERTa and training with two optimizers  \cite{liu-lapata-2019-text}, we obtain two strong baselines BERTS2S and RoBERTaS2S.
T-DMCA is the best model presented by \citet{DBLP:conf/iclr/LiuSPGSKS18} for summarizing long sequence.
HT is the best model presented by \citet{liu-lapata-2019-hierarchical} with the hierarchical Transformer encoder and a flat Transformer decoder.
GraphSum, presented by \citet{li-etal-2020-leveraging}, leverages paragraph-level explicit graph by the graph encoder and decoder, which gives the current best performance on WikiSUM.
We report the best results of  GraphSum with  RoBERTa and the input length is about 2400 tokens.
The last block reports the results of our model BASS with the input lengths of 2400 and 3000. 
Compared with all the baselines, our model BASS achieves great improvements on all the four metrics.
The results demonstrates the effectiveness of our phrase-level semantic graph comparing with other RoBERTa based models, RoBERTaS2S (without graph) and GraphSum (sentence-relation graph). 
Furthermore, the phrase-level semantic graph improves the semantic relevance of the generated summaries and references, as the BERTScore improvements of BASS is obvious.\\
\begin{table}
\begin{tabular}{l|llll}
\hline
Model&R-1&R-2&R-L&BS\\
\hline
\textbf{Full model}&\textbf{42.29}&\textbf{27.19}&\textbf{36.46}&\textbf{30.62}\\
w/o structure&41.86&27.06&36.43&29.84\\
+w/o merging& 41.56&26.61&35.93&29.15\\
\hline
\end{tabular}
\caption{Graph Structure analysis on WikiSUM test set where the input length is 800. \textbf{w/o structure} and \textbf{+w/o merging} refer to remove relations between phrases and further remove phrase merging in graph construction, respectively.}
\label{ablationgraph}
\end{table}
\begin{table}
\centering
\begin{tabular}{l|llll}
\hline
Model&R-1&R-2&R-L&BS\\
\hline
\textbf{Full model}&\textbf{43.40}&\textbf{28.50}&\textbf{37.71}&\textbf{31.64}\\

w/o shortcut &42.50&27.97&37.23&31.10\\

w/o supernode&42.93&28.08&37.42&31.15\\
w/o graph-prop&42.84&28.14&37.42	&31.33\\
w/o graph &42.05&27.00&36.56&29.13\\
\hline
\end{tabular}
\caption{Ablation study on WikiSUM test set where the input length is 1600. \textbf{graph-prop} is the abbreviation of graph-propagation. }
\label{ablation}
\end{table}
\textbf{Results on SDS}\quad Table \ref{table:exp1} shows our experiment results along with other SDS baselines.
Similar to WikiSUM, we also report LexRank, TransS2S, and RoBERTaS2S.
Besides, we report the performance of several other baselines.
ORACLE is the upper-bound of current extrative models.  
Seq2seq is based on LSTM encoder-decoder with attention mechanism \cite{DBLP:journals/corr/BahdanauCB14}.
Pointer and Pointer+cov are pointer-generation  \cite{see-etal-2017-get}  with   and without coverage mechanism, respectively.
FastAbs \cite{DBLP:journals/corr/abs-1805-11080} is an abstractive method by jointly training sentence extraction and compression.
TLM \cite{pilault-etal-2020-extractive} is a recent  long-document summarization method based on language model.
We also report the performances of recent pretrianing-based SOTA text generation  models BART (large) and Peaguasus (base) on BIGPATENT, which both contain a parameter size of  406M .
The last block shows the results of our model, which contains a parameter size of  201M .
The results show that BASS  consistently outperforms RoBERTaS2S, and comparable with current large SOTA models with only half of the parameter size.
This further demonstrates the effectiveness of our graph-augmented model on long-document summarization. 

\subsection{Model Analysis}\label{sec:validate}
\begin{figure}
    \includegraphics[scale=0.5]{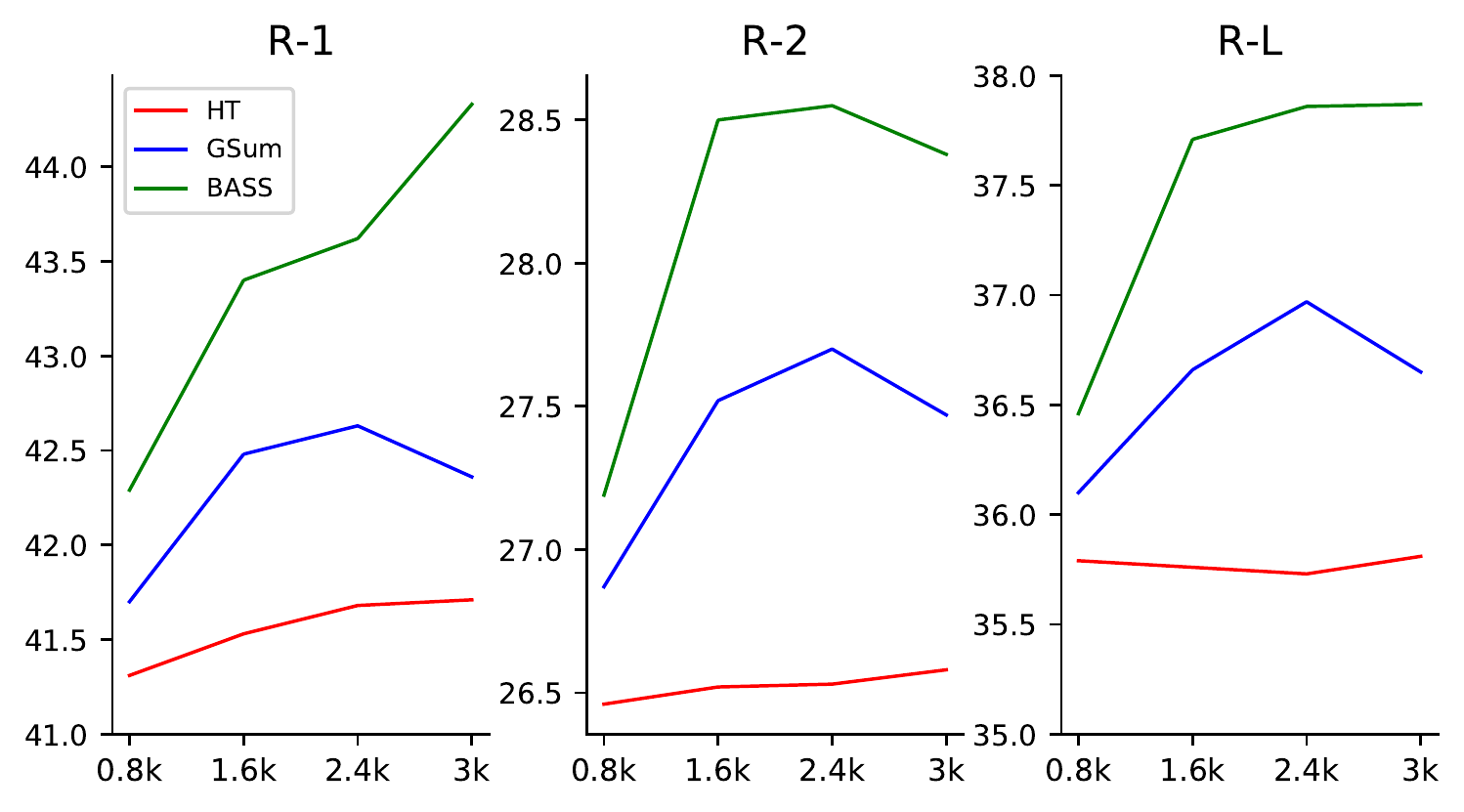}
    \caption{Comparison of HT, GraphSum (GSum in figure), BASS under various length of input tokens.}
    \label{fig:len_comp}
\end{figure}

\begin{figure}
    \includegraphics[scale=0.55]{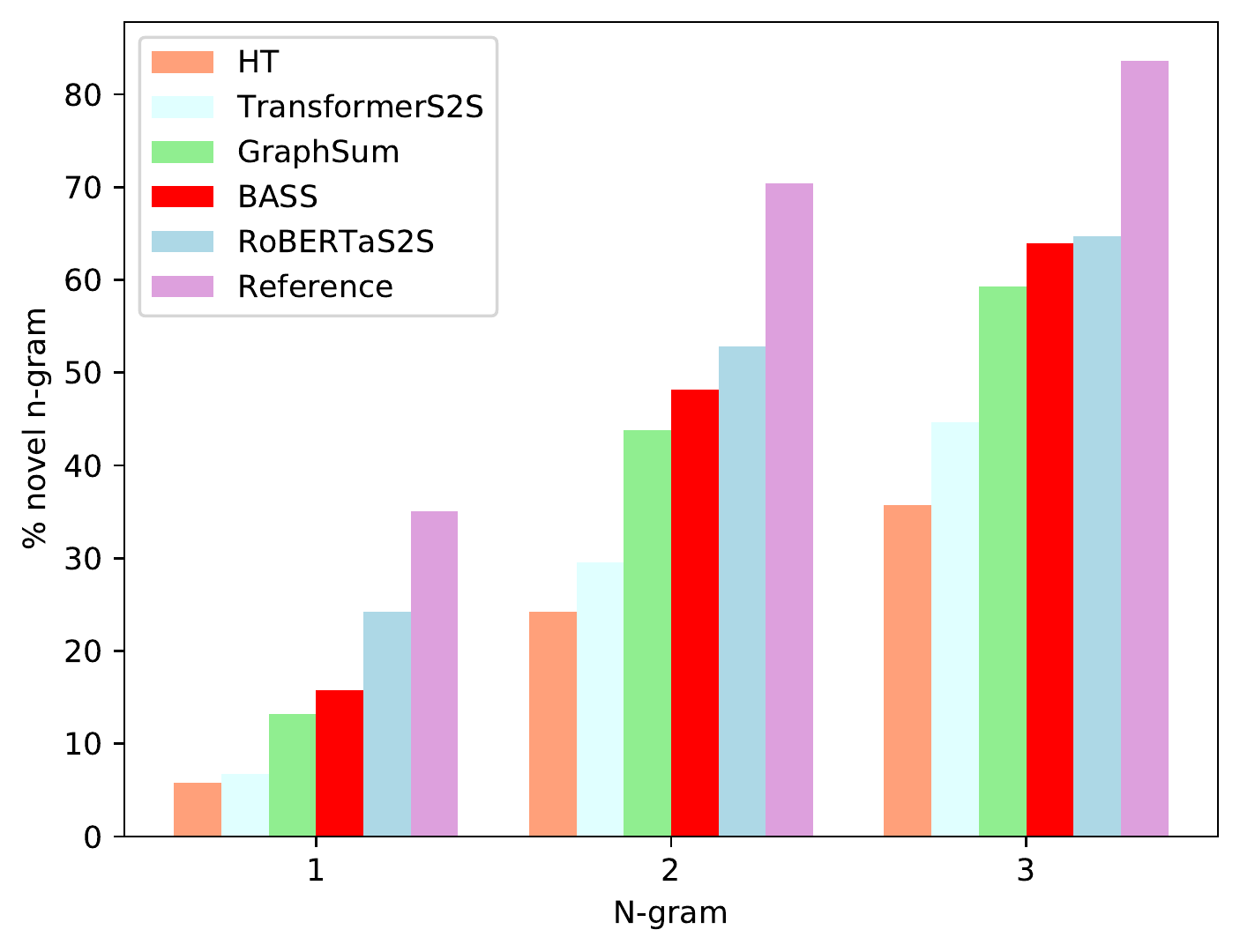}
    \caption{Illustration of novel n-grams in generated summaries form different systems. }
    \label{fig:method_bas}
\end{figure}

For a thorough understanding of BASS, we conduct several experiments on the WikiSUM test set, including the effects of the graph structure and input length. 
We also validate the effectiveness of the graph-augmentation methods in graph encoder and the graph-propagation attention in  graph decoder by ablation studies. \\
\textbf{Graph Structure Analysis} \quad
To analyze how the unified semantic graph benefits summarization learning, we conduct ablation studies on the graph structures.
Illustrated in Table \ref{ablationgraph}, after removing explicit relations between phrases by fully connecting all the nodes, the R-1 metric drops obviously which indicates the relations between phrases improve the informativeness of generated summaries.
After further removing phrase merging, we observe a performance decrease in all the metrics, which indicates the  long-distance relations benefit both the informativeness and fluency of summary.
\\ 
\textbf{Ablation Study}\quad The experimental results of removing supernode and shortcut edges from the unified semantic graph prove the effectiveness of graph augmentation methods in the graph encoder.
Experimental  results without the gaph-propagation attention confirms that the structure of the unified semantic graph is also beneficial for decoding.
Overall, the performance of the model drops the most when removing shortcut edges which indicates the rich potential information is beneficial for summarization.
Finally, after removing all the graph-relevant components, performance dramatically drops on all the metrics.
 \\
\textbf{Length Comparison} \quad According to \citet{DBLP:conf/iclr/LiuSPGSKS18}, input length affects the summarization performance seriously for Seq2Seq models as most of them are not efficient at handling longer sequences.
The basic TransS2S achieves its best performance at the input length of 800, while longer input hurts performance.
Several previous models achieve better performance when utilizing longer sequences.
As illustrated in Figure \ref{fig:len_comp},  the performance of HT remains stable when the input length is longer than 800.
Leveraging the power of sentence-level graph, GraphSum achieves the best performance at 2,400 but its performance begins to decrease when the input length reaches 3000.
Unlike previous methods, ROUGE-1 of BASS  significantly increased in 3000 indicates that the unified semantic graph benefits salient information selection even though the input length is extreme.\\
\textbf{Abastractiveness Analysis}\quad We also study the abstractiveness of BASS and other summarization systems on WikiSUM.
We calculate the average novel n-grams to the source input, which reflects the abstractiveness of a summarization system \cite{see-etal-2017-get}.
Illustrated in Figure \ref{fig:method_bas}, BASS generates more abstract summaries comparing to recent models, GraphSum, HT, and weaker than RoBERTaS2S.
Summarized from observation,  we draw to a conclusion that RoBERTaS2S usually generates context irrelevant contents due to the strong pretrained RoBERTa encoder but a  randomly initialized decoder that relays on the long-text input poorly.
Graph-based decoders of BASS and GraphSum alleviate this phenomenon.

\subsection{Human Evaluation}
In addition to the above automatic evaluations, we also conduct human evaluations to assess the performance of systems.
Because the patent dataset BIGPATENT contains lots of terminologies and requires professional background knowledge for annotators, we select  WikiSUM as the dataset for evaluations.
As Wikipedia entries can be summarized in many different aspects, annotators will naturally favor systems with longer outputs.
Thus we first filter instances that the summaries of different systems are significantly different in lengths and then randomly select 100 instances.
We invite 2 annotators to assess the summaries of different models independently.

Annotators evaluate the overall quality of summaries by ranking them taking into account the following criterias: (1) \textit{Informativeness}: whether the summary conveys important and faithful facts of the input? (2) \textit{Fluency}: whether the summary is fluent, grammatical, and coherent? (3) \textit{Succinctness}: whether the summary is concise and dose not describe too many details? 
Summaries with the same quality get the same order.
All systems get score 2,1,-1,2 for ranking 1,2,3,4 respectively.
The rating of each system is averaged by the scores of all test instances.

The results of our system and the other three strong baselines are shown in Table \ref{tab:human_eval}.
The percentage of rankings and the overall scores are both reported.
Summarized from the results, our model BASS is able to generate higher quality summaries.
Some examples are also shown in the appendix.
Specifically, BASS generates fluent and concise summaries containing more salient content compared with other systems.
The human evaluation results further validate the effectiveness of our semantic graph-based model.
\section{Conclusion and Future Work}
In this paper, we propose to leverage the unified semantic graph to improve the performance of neural abstractive models for long-document summarization and MDS.
We further present a  graph-based encoder-decoder model to improve both the document representation and summary generation process by leveraging the graph structure.
Experiments on both long-document summarization and MDS show that our model outperforms several strong baselines, which demonstrates the effectiveness of our graph-based model and the superiority of the unified semantic graph for long-input abstractive summarization. 
\begin{table}
    \centering
    \begin{tabular}{cccccc}
    \hline
    \textbf{ Model}&1&2&3&4&Rating  \\
    \hline
         TransS2S&0.32&0.14&0.09&0.45&$-0.21^*$\\
         R.B.&0.39&0.22&0.26&0.13&$0.48^*$\\
         G.S.&0.31&0.38&0.20&0.11&$0.58^*$\\
         BASS&0.64&0.16&0.14&0.06&\textbf{1.18}\\
         \hline
    \end{tabular}
    \caption{Ranking results of system summaries by human evaluation. 1 is the best and 4 is the worst. The larger rating denotes better summary quality. R.B. and G.S. are the abbreviations of RoBERTaS2S and GraphSum. * indicates the overall ratings of the corresponding model are significantly (by Welchs t-test with p \textless 0.01) outperformed by BASS.}
    \label{tab:human_eval}
\end{table}
Though remarkable achievements have been made by neural network-based summarization systems, they still do not actually understand languages and semantics.
Incorporating language structures in deep neural networks as prior knowledge is a straightforward and effective way to help summarization systems, as proved by this work and previous works.

\section*{Acknowledgments}
This work was partially supported by National Key R\&D Program of China (No. 2020YFB1406701) and National Natural Science Foundation of China (No. 61876009).


\bibliographystyle{acl_natbib}
\bibliography{anthology,acl2021}
\newpage
\appendix

\section{Graph Construction}
Given a document set with $n$ documents $D = \{d_1,...d_n\}$ and each document $d_i \in D$ contains $k_i$ sentences. 
Algorithm \ref{algorithm} gives the details of constructing the unified semantic graph based on dependency parsing.

We apply CoreNLP for both coreference resolution and dependency parsing.
We first extract coreference chains from every document and merge coreference chains with overlap phrases.
We memorize all the coreference chains in set $C$, where each chain $c=\{p_1,...,p_{k_c}\} \in C$ contains a set of co-referent phrases.
We then parse every sentence in every document into a dependency parsing tree $T_s$.
\begin{algorithm}[htb] 
 \small{
\caption{\textbf{Construct Unified Semantic Graphs}}\label{algorithm}
\KwIn{Documents set $\mathcal D= \{d_1,...,d_n \}$, document $ d_i \in \mathcal D$, $d_i=\{s_1,...,s_{k_i} \}$}
\KwOut{The unified semantic graph $\mathcal{G}$}
\Comment{Coreference Resolution}\\
$\mathcal C \gets \emptyset$\\
\ForEach{$d \in \mathcal D $}{
     $c_d \gets$  COREFERNCE\_RESOLUSION$(d)$ \\
     $\mathcal C \gets$ COREFERNCE\_MERGE($\mathcal C,c_d$)
}
\Comment{Dependency Parsing}\\
$\mathcal T \gets \emptyset$\\
\ForEach {$d \in \mathcal D $}{
\ForEach {$s \in d $}{
 $T_s \gets$ DEPENDENCY\_PARSE($s$)\\
\ $T_s \gets$ IDENTIFY\_NODE\_TYPES($T_s$)\\
\ $T_s \gets$ REMOVE\_PUNCTUATION($T_s$)\\
\ $T_s \gets$ MERGE\_COREF\_PHRASE($T_s, \mathcal C$)\\
\ $T_s \gets$ MERGE\_NODES($T_s$)\\
\ $\mathcal T \gets \mathcal T \bigcup \{T_s\}$
}
}

\Comment{Initialize Graph}\\
$\mathcal G=( \mathcal V, \mathcal E),\mathcal V \gets \emptyset,\mathcal E \gets \emptyset$\\
\ForEach {tree $T=(V_T,E_T) \in  \mathcal T$}{
$ \mathcal V \gets \mathcal V \bigcup \{V_T\} $\\
$ \mathcal E \gets \mathcal E \bigcup \{E_T\} $\\
}
\Comment{Merge Co-referent Nodes}\\
\ForEach{corefernce chain $c \in \mathcal C$}{
$(\mathcal V,\mathcal E) \gets$  MERGE\_PHRASE($c,\mathcal V,\mathcal E$)
}
$\mathcal G \gets ( \mathcal V, \mathcal E)$\\
\Return{$\mathcal G$}
\label{build_graph}}
\end{algorithm}
Afterwords we refines the tree by following operations:
\begin{itemize}
\item IDENTIFY\_NODE\_TYPES: after dependency parsing, each node in the tree is attached with a POS tag. We associate every node with its POS tag for future merging operations.
\item PRUNE\_PUNCTUATION: we remove all the punctuation nodes and their edges.
\item  MERGE\_COREFE\_PHRASE: since a coreference chain contains a set of phrases but a  dependency parsing tree is based on tokens, we first obtain phrases in coreference chains for the future convenience in merging coreferent phrases. For every phrase  $p_i$  in a co-reference chain $c$, we merge the corresponding tokens of $p_i$ to form the target phrase $p_i$ in the tree. The merging operation is carried out by removing edges between the nodes and represent the tokens as a unified node.

\item NODE\_MERGE: after obtaining phrases in coreference chains, we merge other token nodes into concise phrases. This procedure is carried out by traveling every dependency graph in depth-first, and merge the tokens into a phrase if they satisfy the merging conditions. Overall, we merge consecutive tokens that form a complete semantic unit into a phrases.
\end{itemize}
After we extract all the phrases, we merge all the same phrases and phrases in the same coreference chain by MERGE\_PHRASE and return the final semantic graph.

\section{Case Study}

We select several cases from human evaluation and demonstrate them to show the overall quality of systems.
In each table, there are four blocks present the  input article (\textbf{Article}), the reference summary (\textbf{Reference Summary}), the output summary of a strong baseline GraphSum (\textbf{Baseline}) and  the output summary of our model BASS (\textbf{BASS}),  separately.
The original input article is the concatenation of  several document paragraphs by the ``\(\vert \vert\)" symbol containing 1600 tokens in maximum.
We only show the salient part of the input article due to the paragraph constraints.
\colorbox{yellow}{Spans in highlight} indicate the salient contents.
\color{red} Spans in  red \color{black} indicate the unfaithful content, irrelevant content or repeats a system generated.
The case in Table \ref{tab:colleen_coyne} describes an American ice hockey player ``Colleen Coyne".
The important fact, ``won a gold medal at the 1998 winter Olympics", is well captured by BASS, however, the baseline model only mentions she ``was a member" neglecting the substantial achievement.
The case in Table \ref{tab:havana_quartet} introduces the play ``Colleen Coyne'' which based on the four novels of ``Leonardo Padura'' is difficult to summarize, as the relation between ``Colleen Coyne'', ``Leonardo Padura'' and the name list of the four novels cross different documents and a long-span.
The baseline model confuses with the name of stars and fails to list the names of four books.
The \textbf{Reference Summary} in Table \ref{tab:cetacean_intelligence} is not informative enough to give a precise description of what is ``Cetacean Intelligence".
Though BASS does not introduce the definition of ``Cetacean", it clearly describes the categories of  ``Cetacean Intelligence" which is more essential to the topic.
In Table \ref{tab:james_tolbert},  BASS and Baseline generate summaries with similar content, but  BASS provides more details such as, ``right-handed'',  distributed in different documents.
In the case describing Broadcast, in Table \ref{tab:boadcast}, while the Baseline generates irrelevant titles of editors, BASS describes essential characters of the magazine.
Though all the models apply trigram-block   to avoid repeats, Table \ref{tab:adams_building} shows that sometimes the Baseline still generates repeated n-grams while this seldomly happens on BASS.
\begin{table*}
    \centering
    \begin{tabular}{|p{15.5cm}|}
         \hline
        
        \textbf{ Article:}\colorbox{yellow}{colleen coyne} is a graduate of the university of new hampshire. \colorbox{yellow}{an ice hockey player}, she represented \colorbox{yellow}{the united states}, as a defenseman, on 6 nat ...  \(\vert \vert\)  ... \colorbox{yellow}{colleen coyne} was one of the trailblazing women who won gold at the \colorbox{yellow}{1998 olympic winter games}. ...\(\vert \vert\)...history and heroes:george nagobads, mike ilitch's pizza \& \colorbox{yellow}{colleen coyne} of \colorbox{yellow}{the 1998 u.s. women's olympic team}...  
        \\[6pt]
         \hline
          
        \textbf{Reference Summary:} colleen m. coyne (born september 19, 1971 ) is an american ice hockey player. she won a gold medal at the 1998 winter olympics.
        \\[6pt]
        \hline

        \textbf{Baseline:} colleen coyne (born november 3, 1974) is an american former ice hockey player.  she was a member of the united states women 's national ice hockey team at the 1998 winter olympics.
        \\[6pt]
         \hline
         
          \textbf{BASS:} colleen coyne is an american ice hockey player.  \colorbox{yellow}{she won a gold meda at the 1998 winter olympics.}  \colorbox{yellow}{ 1998 winter olympics.}
          \\[6pt]        \hline
    \end{tabular}
    \caption{Colleen Coyne}
    \label{tab:colleen_coyne}

    \begin{tabular}{|p{15.5cm}|}
         \hline
        
        \textbf{Article:}\colorbox{yellow}{``havana quartet''}, based on \colorbox{yellow}{the popular book series} by cuban novelist \colorbox{yellow}{leonardo padura}, follows a hard-drinking, romantic cuban police detective mario conde -lrb- played by banderas -rrb- who longs to be a writer but settled for a job as a detective.\(\vert \vert\)\colorbox{yellow}{``havana quartet''} follows hard-drinking, romantic cuban police detective mario conde -lrb- banderas -rrb- who longs to be a writer but settled for a job as a detective.   \(\vert \vert\)  known cuban playwright eduardo machado -lrb- starz 's magic city -rrb-, who lives in the u.s., is the writer on the project, based on \colorbox{yellow}{the popular four-book series of detective novels havana blue, havana gold, havana red} \colorbox{yellow}{and havana black} by another cuban writer that is well known internationally, novelist \colorbox{yellow}{leonardo padura}. 
        \\[6pt]
         \hline
        \textbf{Reference Summary:} havana quartet is an upcoming american television drama series starring antonio banderas. the series is based on \colorbox{yellow}{four detective novels} by cuban author\colorbox{yellow}{ leonardo padura} , \colorbox{yellow}{havana blue , havana gold , havana red and havana black}.
        \\[6pt]
          \hline
         
        \textbf{Baseline:} havana quartet is an upcoming american television drama series based on the novel series of the same name by cuban novelist leonardo padura the series stars \color{red}{antonio banderas, eduardo machado and eduardo machad}.
        \\[6pt]
         \hline
          \textbf{BASS:}  havana quartet -lrb- spanish : la gazeta de havana -rrb- is an upcoming american television series baed on \colorbox{yellow}{leonardo padura 's novels havana blue , havana gold , havana red and havana black.}
          \\[6pt]        \hline
    \end{tabular}
    \caption{Havana Quartet}
    \label{tab:havana_quartet}
    \begin{tabular}{|p{15.5cm}|}
         \hline
        
        \textbf{Article:}in the past three decades, new research has revealed that dolphin brains are not only large but are extremely complex in their organization, forming the neurobiological basis for the considerable intelligence and socio-cultural characteristics of dolphins and their relatives, the whales. dolphin brains are larger than expected for their body size and contain numerous features associated with sophisticated perceptual and \colorbox{yellow}{cognitive abilities, echolocation, communication, cooperation, and dynamic social networks}, cultural transmission of learned behaviors \colorbox{yellow}{, and self-awareness}, to name a few.
        \\[6pt]
         \hline
        
       \textbf{Reference Summary:} cetacean intelligence refers to the cognitive capabilities of the cetacea order of mammals. this order includes whales, porpoises, and dolphins.
        \\[6pt]
        \hline
          
        \textbf{Baseline:} cetacean intelligence ( cetacean intelligence ) refers to the ability of cetaceans to communicate with other animals. cetacean intelligence is an object of fascination among scientists.
        \\[6pt]
         \hline
          
          \textbf{BASS:}cetacean intelligence refers to the ability of cetaceans to communicate with other animals. \colorbox{yellow}{cetacans have a wide range of social and cognitive abilities, including echolocation, communication,} \colorbox{yellow}{cooperation and dynamic social networks, cultural transmission of learned behaviors,and } \colorbox{yellow}{self-awareness}.
          \\[6pt]        \hline
    \end{tabular}
    
    \caption{Cetacean Intelligence}
    \label{tab:cetacean_intelligence}
    
\end{table*}

\begin{table*}
    \centering
    
    \begin{tabular}{|p{15.5cm}|}
         \hline
        
        \textbf{Article:}\(\vert \vert\)\colorbox{yellow}{james tolbert hearn} was born on april 11 , 1921 , in atlanta , georgia , the sixth and final child born to raymond and pearl ( adams ) hearn  \(\vert \vert\) i certainly had no idea of ever being a pitcher . i had never pitched as a kid , " said 13-year big-league veteran big jim hearn , who won 109 games for the\(\vert \vert\) \colorbox{yellow} {st. louis cardinals , new york giants , and philadelphia phillies from 1947 through 1959 }.\(\vert \vert\) \quad \colorbox{yellow} {jim hearn } a \(\vert \vert\)\colorbox{yellow}{right-handed pitcher} who helped propel the new york giants to victory ...
        \\[6pt]        \hline
        
       \textbf{Reference Summary:} james tolbert hearn ( april 11 , 1921 -- june 10 , 1998 ) was an american right-handed pitcher in major league baseball for 13 seasons ( 1947 -- 59 ) .   a right-hander , he stood 6 feet 3 inches ( 1.91 m ) tall and weighed 205 pounds ( 93 kg ) .
        \\[6pt]
        \hline
         
        \textbf{Baseline:} james tolbert hearn ( april 11 , 1921 -- june 10 , 1998 ) was an american professional baseball pitcher .  he played 13 seasons in major league baseball ( mlb ) for the st. louis cardinals , new york giants , and philadelphia phillies .
        \\[6pt]
         \hline
          
          \textbf{BASS:}james tolbert hearn ( april 11, 1921 -- june 10, 1998 ) was an american \colorbox{yellow}{right-handed} pitcher in major league baseball who played for the st. louis cardinals, new york giants , \colorbox{yellow}{and philadelphia phillies from 1947 through 1959} .
          \\[6pt]        \hline
    \end{tabular}
    \caption{James Tolbert Hearn}
    \label{tab:james_tolbert}
    \begin{tabular}{|p{15.5cm}|}
         \hline
        
        \textbf{Article:}\colorbox{yellow}{broadcast} ( magazine )  content s.v.p. / chief content officer derek t. dingle s.v.p. / executive editor-at-large alfred a. edmond jr. . v.p. / director of ...\(\vert \vert\)\colorbox{yellow}{broadcast}, \colorbox{yellow}{a weekly magazine} for the uk television and radio industry , ...\(\vert \vert\) broadcast is the ideal supplement for those working in the uk broadcasting industry , offering up-to-date industry news from the most experienced and \colorbox{yellow}{authoritative voices} in broadcasting \(\vert \vert\)as a broadcast subscriber , you will benefit from the uk tv industry's most knowledgeable source of industry news and \colorbox{yellow}{analysis on the successful people , programmes and companies}.from knowing what commissioners are looking for , to the latest funding opportunities ,  
        \\[6pt]        \hline
        
       \textbf{Reference Summary:} broadcast is a weekly magazine for the united kingdom television and radio industry .   it covers a wide range of news and issues affecting the professional broadcast market in the uk .   broadcast has regular weekly sections covering news ...
        \\[6pt]
        \hline
         
        \textbf{Baseline:} broadcast is a british monthly magazine covering the uk television industry .  it is published by content \color{red}.v.p. . \color{black} the magazine was established in 1931 and is edited by alfred a. edmond , \color{red}jr. .  v.p.
        \\[6pt]
         \hline
          
          \textbf{BASS:}broadcast is a monthly magazine covering the uk television industry .  the magazine was first published in 1931 .  it is the uk 's most authoritative voices of industry news and analysis \colorbox{yellow}{on the successful people , programmes and companies} .
          \\[6pt]        \hline
    \end{tabular}
    \caption{Broadcast}
    \label{tab:boadcast}
     \begin{tabular}{|p{15.5cm}|}
         \hline
        
        \textbf{Article:}dams building ( sault ste. marie , michigan )  npgallery allows you to search the national register information system a database of over 90,000 historic buildings ,... ( added \colorbox{yellow}{2010} - - \# 10000218 ) also known as central savings bank building \colorbox{yellow}{418 ashmun st. , sault ste. marie} \(\vert \vert\)  for those of you who are interested in working with data in a gis environment , there is more information at national register of historic places program research page .
        \\[6pt]        \hline
        
       \textbf{Reference Summary:} the adams building , also known as the central savings bank building , was built as a commercial and office building located at 418 ashmun street in sault ste. marie , michigan .   ...   .it was listed on the national register of historic places in 2010 .
        \\[6pt]
        \hline
         
        \textbf{Baseline:} the adams building , also known as the central savings bank \color{red}building building \color{black}, is a building located at 418 ashmun street in sault ste. marie , michigan .  it was listed on the national register of historic places in 2010 .
        \\[6pt]
         \hline
          
          \textbf{BASS:}the adams building , also known as the central savings bank building , is a commercial building located at 418 ashmun street in sault ste. marie , michigan .  it was listed on the national register of historic places in 2010 .
          \\[6pt]        \hline
    \end{tabular}
    \caption{The Adams Building}
    \label{tab:adams_building}
\end{table*}



\end{document}